\documentclass[10pt]{article}

\usepackage{microtype}
\usepackage{graphicx}
\usepackage{subfigure}
\usepackage{booktabs}

\usepackage{xargs}
\usepackage[colorinlistoftodos,prependcaption,textsize=tiny]{todonotes}
\usepackage{amsmath}
\usepackage{amssymb}
\usepackage{epstopdf}
\usepackage{gensymb}
\usepackage{url}
\usepackage{gensymb}
\usepackage{lmodern}
\usepackage{wrapfig}
\usepackage{appendix}
\usepackage{hyperref}
\usepackage{stmaryrd}
\usepackage{enumitem}
\usepackage{float}
\usepackage{caption}
\usepackage{tikz-qtree}
\usepackage{xcolor}

\usepackage[accepted]{icml2018}
\usepackage{icml2018}
\floatname{algorithm}{Procedure}

\usetikzlibrary{decorations.pathmorphing}
\usetikzlibrary{decorations.markings}
\usetikzlibrary{shapes.geometric}
\usetikzlibrary{arrows}
\usetikzlibrary{shapes.arrows, shapes.callouts, calc}
\newcommand{\midarrow}{\tikz \draw[-triangle 90] (0,0) -- +(.1,0);}
\tikzset{snake it/.style={decorate, decoration=snake},
  very thick/.style={line width=2.6pt}, thick/.style={line width=1.2pt}}
\icmltitlerunning{Exploring Hierarchy-Aware Inverse Reinforcement Learning}

\begin{document}

\twocolumn[
\icmltitle{Exploring Hierarchy-Aware Inverse Reinforcement Learning}

\begin{icmlauthorlist}
  \icmlauthor{Chris Cundy}{ucb}
  \icmlauthor{Daniel Filan}{ucb}
\end{icmlauthorlist}
\icmlaffiliation{ucb}{Department of Electrical Engineering and Computer Science, University of California Berkeley, Berkeley, CA, 94720, USA}

\icmlcorrespondingauthor{Chris J.\ Cundy}{chris.j.cundy@gmail.com}

\vskip 0.3in
]

\printAffiliationsAndNotice{}

\begin{abstract}
  We introduce a new generative model for human planning under the Bayesian
  Inverse Reinforcement Learning (BIRL) framework which takes into account the
  fact that humans often plan using hierarchical strategies. We describe the
  Bayesian Inverse Hierarchical RL (BIHRL) algorithm for inferring the values of
  hierarchical planners, and use an illustrative toy model to show that BIHRL retains
  accuracy where standard BIRL fails. Furthermore, BIHRL is able to accurately
  predict the goals of `Wikispeedia' game players, with inclusion of
  hierarchical structure in the model resulting in a large boost in accuracy. We
  show that BIHRL is able to significantly outperform BIRL even when we only
  have a weak prior on the hierarchical structure of the plans available to the
  agent, and discuss the significant challenges that remain for scaling up this framework to more realistic settings.\footnote[2]{Our implementation of the algorithm can be found at https://github.com/C-J-Cundy}

\end{abstract}

\section{Introduction}


As Reinforcement Learning (RL) algorithms have become more and more capable, we are increasingly aware of
the limitations of how we specify their goals. While these goals can be hand-crafted
for simple environments, this approach requires expert knowledge in the domain.
If we are to eventually use AI to perform tasks that are beyond human abilities
(e.g.\ `plan a city'), we have to develop a more robust method of goal specification.
Our algorithms would ideally be able to learn what goals they should pursue by
inferring human preferences: this is often known as \emph{value learning}, or
\emph{preference elicitation}.

A leading approach to value learning from observed human actions
is inverse optimal control \cite{Kalman1960} or inverse reinforcement learning (IRL),
formalised by \citet{Ng2000} and
\citet{Abbeel2004}.
In IRL we treat human behaviour as planning in a Markov decision process (MDP)
and aim to find a reward function that explains observed trajectories of human agents.

While we may naively assume that human beings always act perfectly to
achieve their goals (the `principle of revealed preference' in
economics \citep{Samuelson1938}), human behaviour often violates this
assumption.  In general, people make choices that they admit are suboptimal, due
to a variety of biases including lack of willpower, inconsistent
time preferences, and lack of perfect foresight.  Therefore, a more accurate
inference of `true' preferences must take typical human irrationality into
account. Although initial approaches to IRL followed this implicit assumption of rationality
of the demonstrating expert, the more recent Bayesian IRL framework
\citep{Ramachandran2007} makes it straightforward to include more realistic
models of human behaviour. Previous work in this area has modelled human
actions as attempting to maximise their utility subject to constraints such as
limited knowledge \citep{Baker2014} or inconsistent time
preferences \citep{Evans2015}.

However, to our knowledge no previous work has considered what we believe to be
a key feature of human planning: a tendency to structure our decision-making in
a hierarchical fashion. Instead of evaluating each individual action in terms of
the rewards which we expect to obtain from all subsequent actions, humans tend
to simplify their planning by considering sub-problems and choosing between
known methods to solve these problems. For example, when navigating across a
city we might choose between existing skills of walking, taking a taxi or public
transport. We do not choose between all the trajectories that we could
physically perform.

If we simply apply existing algorithms to observations of humans who plan in
this way, we will fail to infer correct preferences, running the risk of
accidentally inferring pathologically wrong values in order to explain the
hierarchically-generated plans.

Our key contributions are as follows:

\begin{itemize}
  \item{We introduce a generative model of human decisions as resulting from hierarchical planning,
      which uses both primitive actions and extended options comprised of sequences of actions.}
    \item{We discuss the theoretical justification for considering
        such a model and introduce a simple algorithm for inference with hierarchically-generated trajectories.}
    \item{Evaluating our model on
        trajectories of players of the `Wikispeedia' game shows us that incorporating
        hierarchical structure gives us a sizeable boost in goal prediction accuracy compared to
        standard Bayesian IRL.}
    \item{Finally, we discuss how our inference procedure can be extended to
        jointly infer options and preferences, and show that our performance
        advantage over BIRL is retained even when we don't know what the precise
        hierarchical structure of the agent is.}
\end{itemize}
    
\section{Our Model}
\label{sec:our_model}
An MDP is a tuple \((\mathcal{S}, \mathcal{A}, T, R, \gamma)\) consisting of a set of
states \(\mathcal{S}\) and actions \(\mathcal{A}\), a transition function \(T\), reward function \(R\), and discount rate \(\gamma\), following the usual definition in
e.g.\ \citet{Sutton2012}. In IRL we are given an MDP without \(R\) and aim to
recover the reward from an observed trajectory of an agent's actions and entered states at each timestep \(\mathcal{T}_a = (s_0, a_0), (s_1, a_1), \ldots \). (We need to include the states as actions do not uniquely map to states in a stochastic MDP). We can typically extend the inference over multiple observed trajectories.

We describe the behaviour of an agent in an MDP by a stochastic policy \(\pi\). We write the optimal policy as \(\pi^*\), with corresponding Q-function
\(Q^*\). 
Human planning is commonly modelled as being Boltzmann-rational: that is, satisfying $\pi (s,a) \propto \exp(\beta Q^*(s,a)) $ for a fixed parameter $\beta$. Boltzmann-policies can also be self-consistent, so that the value-function is computed taking into account the Boltzmann-rational policy. This gives a policy $\pi(s,a) \propto \exp(\beta Q^\odot(s,a)) $, where \(Q^\odot\) is the\footnote{In general there is no unique consistent Boltzmann-policy \citep{Asadi2016}. In practice we have not noticed any problems arising from this non-uniqueness.} Q-value under this same Boltzmann-rational policy. The parameter $\beta$ can be increased or decreased to model more or less rational humans, respectively.

One method for describing the behaviour of agents that plan hierarchically is
the options framework, comprehensively described by \citet{Sutton2015}. An
option \(o\) consists of a policy \(\pi_o\), an initiation set \(\tau \subseteq
\mathcal{S}\), and a termination function \(\alpha: \mathcal{S} \rightarrow [0,
1]\).  
The initiation set \(\tau \subseteq \mathcal{S}\) gives the states where the agent may activate the policy, thereafter following the policy \(\pi_o\). At each state \(s\) the policy enters, the termination function \(\alpha(s)\) gives the probability that the option terminates, after which the agent no longer follows \(\pi_o\).
These parameters define an exit distribution \(P^o(s,s^\prime)\) giving the
probability that the option \(o\), if initiated in state \(s\), will terminate
in state \(s^\prime\), and a reward function \(r^o(s)\) giving the expected
reward for activating option \(o\) in state \(s\). For a given state-action sequence \(\mathcal{T}_a\), we can further consider the consistent-exit distribution
\(P^{oc}(s,s^\prime, \mathcal{T}_a)\). This gives the probability that taking
the option \(o\) in state \(s\) results in the option's policy giving the exact state-action trajectory in \(\mathcal{T}_a\), terminating
in state \(s^\prime\). An action \(a\) in a state \(s\) in an MDP can be described as a
degenerate option where \(\pi_o(a,s)=1\), \(\tau = \{s\}\), and \(\alpha(s_1) =
1\) if \(T(s_1, s, a) \neq 0\). Our use of the term `option' includes these
`atomic' actions as a special case.

Thus the key features of our model are as follows:

\begin{itemize}
\item{The human has an available set of options \(\omega\), which include
    options with a policy that terminates after one action, i.e.\ the
    standard actions in the MDP.}
\item{The human chooses between options \(o \in \omega\) with a stochastic policy,
    \(\pi(s,o) \propto \exp(\beta Q^\odot(s, o))\) for a fixed parameter \(\beta\).}
\item{We do not observe the sequence of \emph{options} that the agent executes: we only observe the
    sequence of states and \emph{actions} \(\mathcal{T}_a\), that the agent executes,
    some of which may have been executed as part of a compound option. We denote the unobserved state-\emph{option} trajectory by \(\mathcal{T}_o\).}
\end{itemize} 
 
A key feature of our model is the inclusion of Boltzmann-rational decisions over extended options as well as single actions. We believe that this feature is important for accurate modelling of human preferences, after considering the common everyday situations where the human has options that are well-suited to
solving problems, but are not optimal. The human might take those options instead of
explicitly computing the optimal policy because they have a limited ability to
optimally plan. For instance, if they wish to get across the city, they might choose
between a taxi and walking, as those skills have served them well in the past. They
might not even consider asking to borrow a friend's bicycle, even if this might
be the fastest method, and certainly within their abilities. We wouldn't want our
preference inference algorithm to conclude that the human prefers sitting in
 taxis because they chose to do that over taking the optimal policy. 
 For an overview of the psychology and neuroscience literature on the importance of hierarchy in human planning and the neural basis thereof, see \citet{botvinick2009hierarchically}.

\section{Related Work}
\subsection{Boltzmann-rationality}
The Boltzmann-rationality model of human behaviour is one of the simplest variations on the naive assumption that humans are completely rational, and has a long history in the literature. While it violates certain
assumptions of how agents should act, such as the principle of independence of
irrelevant alternatives introduced by \citet{Debreu1960}, in practice the model
has found widespread use in explaining how people make
bets \citep{Rieskamp2008}; in modelling the attention of people looking at
adverts \citep{Yang2015}; and understanding the decisions taken in the brain itself \citep{Glascher2010}.

Previous work \citep{Ortega2013} has shown how a modified Boltzmann-policy can
arise from modelling bounded agents as they trade off gains in utility against
expending energy to transform their prior probability distributions into
posterior distributions (quantified as a regularisation on the relative entropy
between the two distributions). Under this framework, a Boltzmann-policy is the
optimal policy for an agent which starts out indifferent to its actions, and can
spend an amount of energy characterised by \(\beta\) on investigating which
actions are likely to give it high reward. Seen through this lens, the
Boltzmann-rational human agent has a certain theoretical justification, in
addition to being commonly used in practice.

\subsection{Incorporating human decision-making in IRL}
Initial work on inverse reinforcement learning \citep{Ng2000} did not
discuss the procedure the human used to generate the policy and so implicitly assumed optimality of the human policy. Contemporary work in IRL tends to build on one of two frameworks: Maximum Entropy IRL, introduced by \citet{Ziebart2008}; or
Bayesian IRL (BIRL), introduced by \citet{Ramachandran2007}. For the present work,
we work within the Bayesian IRL framework due to its conceptual simplicity and
straightforward inversion of planning to inference.  Recent work has also built
on BIRL to incorporate non-optimal human behaviour, such as
inconsistent time preferences \citep{Evans2015} or limited
knowledge \citep{Baker2014}.

The most closely related work is by \citet{Nakahashi2016}, who assume
that humans attempt to fulfill a set of goals, which may consist of
subgoals. A Bayesian method is then used to find which parts of the observed
trajectory correspond to fulfilling each goal/subgoal.
While this goal/subgoal setting seems a reasonable assumption for many of the
trajectories, an arbitrarily parameterised reward function can more flexibly model a wider variety of tasks, requiring less domain-specific knowledge. Secondly, their work assumes an inherent hierarchical structure of tasks, whilst our approach assumes that human planners impose this structure as a shortcut for efficient planning, possibly leading to hierarchically optimal but globally suboptimal trajectories.


\section{Taxi-Driver Environment}
\label{sec:taxi-driver-description}
The taxi driver environment was first introduced by \citet{Dietterich2000} as
an example of a task that is particularly amenable to hierarchical reinforcement
learning (HRL) methods. It is a useful running example to describe the mechanics of
hierarchical planning.

\begin{figure}
  \centering
  \begin{tikzpicture}[every node/.style={minimum size=1cm-\pgflinewidth, outer sep=0pt}]
  \node[fill=black!15] at (0.5 ,4.5) {};
  \node[fill=black!15] at (1.5,4.5) {};
  \node[fill=black!15] at (2.5,4.5) {};
  \node[fill=black!15] at (3.5,4.5) {};
  \node[fill=black!15] at (4.5,4.5) {};
  \node[fill=black!15] at (0.5,3.5) {};
  \node[fill=black!15] at (1.5,3.5) {};
  \node[fill=black!15] at (2.5,3.5) {};
  \node[fill=black!15] at (3.5,3.5) {};
  \node[fill=black!15] at (4.5,3.5) {};
  \node[fill=black!15] at (0.5,0.5) {};
  \node[fill=black!15] at (0.5,1.5) {};
  \node[fill=black!15] at (3.5,0.5) {};
  \node[fill=black!15] at (3.5,1.5) {};
  \node[fill=black!15] at (4.5,0.5) {};
  \node[fill=black!15] at (4.5,1.5) {};    
    \draw[step=1.0,black, thick] (0,0) grid (5,5);
    \node at (1.5, 4.5) {\texttt{R}};
    \node at (0.5, 4.5) {\texttt{R}\(_1\)};    
    \node at (4.5, 4.5) {\texttt{G}};    
    \node at (3.5, 0.5) {\texttt{B}};
    \node at (4.5, 0.5) {\texttt{B}\(_1\)};        
    \node at (0.5, 0.5) {\texttt{Y}};
      \draw [very thick] (1,0) -- (1,2);
    \draw [very thick] (2,5) -- (2,3);
    \draw [very thick] (3,0) -- (3,2);
  \begin{scope}[orange!50, very thick, every node/.style={sloped,allow upside down}]    
    \draw (2.3,1.5) -> node {\midarrow} (2.3,2.2) -> node {\midarrow} (1.5, 2.2)
    -> node {\midarrow} (0.5,2.2) -> node {\midarrow} (0.5, 3.5) -> node {\midarrow}
    (0.5, 4.3) -> node {\midarrow} (1.5, 4.3) -> node {\midarrow} (1.5, 3.5) -> node
    {\midarrow} (1.5, 2.8) -> node {\midarrow} (2.5, 2.8) -> node {\midarrow} (3.5,
    2.8) -> node {\midarrow} (4.5, 2.8) -> node {\midarrow} (4.5, 1.5) -> node
    {\midarrow} (4.5, 0.8) -> node {\midarrow} (3.5, 0.8);
  \end{scope}
  \begin{scope}[very thick, dashed, blue!50, every node/.style={sloped,allow
      upside down}] \draw (2.7,1.5) -> node {\midarrow} (2.7,2.4) -> node
    {\midarrow} (1.5, 2.4) -> node {\midarrow} (0.7,2.4) -> node {\midarrow}
    (0.7, 3.5) -> node {\midarrow} (0.7, 4.1) -> node {\midarrow} (1.2, 4.1) ->
    node {\midarrow} (1.2, 3.5) -> node {\midarrow} (1.2, 2.6) -> node
    {\midarrow} (2.3, 2.6) -> node {\midarrow} (3.3, 2.6) -> node {\midarrow}
    (4.3, 2.6) -> node {\midarrow} (4.3, 3.5) ->
    node {\midarrow} (4.3, 4.5);
  \end{scope}
  \end{tikzpicture}
  \caption{The modified taxi-driver situation considered here. Two trajectories shown are
    drawn from an agent that has hierarchical options \texttt{go to R\(_1\)} and \texttt{go to B\(_1\)}.
    In both trajectories the passenger starts at \texttt{R}, while the destination is
    \texttt{B} in the first and \texttt{G} in the second. Greyed-out cells represent destinations
    of the options in the uniform prior over option-sets used in section \ref{sec:priors}.}
  \label{fig:prob2}
\end{figure}
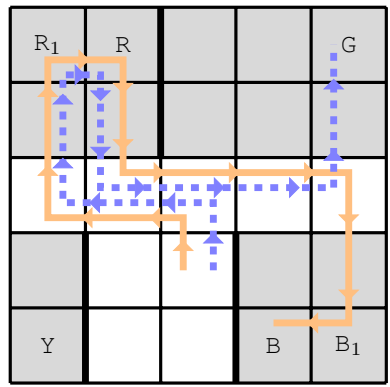
The problem consists of a 5\(\times\)5 gridworld, depicted in
figure \ref{fig:prob2}, with four special landmarks, labelled \texttt{R},
\texttt{G}, \texttt{B} and \texttt{Y}. An agent (the `taxi driver') moves in
this world, starting at a random cell. Additionally, there is a passenger who
initially starts from one of the landmark cells, with a randomly chosen landmark as
their destination. The driver has six different actions: as well as moving in
the cardinal directions with actions \texttt{N}, \texttt{E}, \texttt{S},
\texttt{W}, they can also attempt to \texttt{Pickup} or \texttt{Putdown} the
passenger. The environment gives rewards of \(-1\) on any movement action (attempts
to move into walls or outside the grid fail with no additional penalty), \(-10\) on
unsuccessful attempts to \texttt{Pickup} or \texttt{Putdown}, and \(+20\) on
successfully putting the passenger down at their destination, at which point the
episode terminates. The state consists of the grid coordinate, the location of
the passenger (either at one of the four landmarks or in the taxi), and the desired
destination, giving \(5\times 5\times 5\times 4 = 500\) possible states.

When presented in previous work, the taxi driver is usually equipped with
hierarchical options, such as \texttt{Go to x}, where \texttt{x} is any of
\texttt{R}, \texttt{G}, \texttt{B}, or \texttt{Y} and the environment is used to
show how these allow the problem to be solved faster than without imposing this
structure. Of course, it is somewhat to be expected that an agent will do well if it is provided with options that are exact sub-components of the optimal policy \(\pi^*\). We wish to consider the more realistic setting where the taxi driver
has skills that are well-suited to the task at hand, but not
optimal, i.e.\ they are not exact sub-components of the optimal policy \(\pi^*\), although they are generally much more useful than random policies. Perhaps the human knows how to get to their place of work which is
located in a cell to the right of \texttt{B}, so finds it easier to drive to
\texttt{B} by first going to their place of work, then going west
to \texttt{B}.

Since our aim is to perform IRL in the environment, we consider the variant of the taxi-driver case with a partially observed reward function. We know that the reward is as described above, except that up to five
cells have reward \(0\) to enter (instead of \(-1\) in the standard formulation). We can imagine
this reward as modelling some areas with little traffic, or areas that
the driver enjoys driving along to get to the destination. This means
that we are considering \(\theta\) which are drawn from a finite set with approximately 6.7 million possible reward functions, parameterised by five coordinates giving the locations of the free-to-enter cells.




\section{Bayesian Description}
Given a human
state-action trajectory \(\mathcal{T}_a\) and a set of possible options \(\omega\), we wish
to compute the posterior distribution over a particular parameterisation of the reward
function \(\theta\). In the taxi driver example, \(\mathcal{T}_a\) corresponds to the
sequence of observed actions \texttt{N, E, W}, etc;  \(\omega\) is a set
consisting of concrete actions \texttt{N}, \texttt{S} \ldots, along with some
extended options such as \texttt{Go to B\(_1\)}.

In principle, there is no reason why we cannot consider options consisting of
any stochastic policy, but in order to simplify the experiments we choose to
consider either options with deterministic policies, or options which are
themselves Boltzmann-rational with parameter \(\beta_o > \beta\), where
\(\beta\) is the rationality parameter for the agent's planning over top-level
options. This mirrors the everyday experience of having a set of well-honed
skills that we can count on to give us the outcome we expect.  We choose this
model as we feel it combines being able to plan at
different levels of abstraction (modelled with the availability of multi-action
options) with the limited resources available to plan modelled by the
Boltzmann-rationality \citep{Ortega2013}.

Our inference problem is given by
\begin{equation*} P(\theta | \mathcal{T}_a,\beta, \omega) =
\frac{P(\mathcal{T}_a|\beta, \omega, \theta)P( \theta)}{
P(\mathcal{T}_a| \beta, \omega)}.
\end{equation*} Each observed state-\emph{action} trajectory \(\mathcal{T}_a\)
could have been produced by several state-\emph{option} trajectories
\(\mathcal{T}_{o,i}\), indexed by \(i\). For example, in the taxi-driver case,
we don't know if the driver navigating to \texttt{B\(_1\)} is due to
the driver executing a series of atomic options (\texttt{North},
\texttt{West}, \(\ldots\)), or by executing the single compound option
\texttt{Go to B\(_1\)}.
So we express \(P(\mathcal{T}_a| \beta, \omega)\) in terms of the unobserved option-trajectories \(\mathcal{T}_{oi}\) with \ \(P(\mathcal{T}_a| \beta, \omega) = \sum_i
P(\mathcal{T}_a|\mathcal{T}_{o,i})P(\mathcal{T}_{o,i}| \beta,
\omega)\).\footnote{\(P(\mathcal{T}_a|\mathcal{T}_{o,i})\)
might be less than 1 if the option follows a stochastic policy, e.g.\
an option which itself has a Boltzmann-policy.} Then:
\begin{equation*}
\label{eq:1} P(\theta | \mathcal{T}_a, \beta, \omega) = \frac{\sum_i
P(\mathcal{T}_a|\mathcal{T}_{o,i})P(\mathcal{T}_{o,i}| \beta, \omega, \theta)P(
\theta)}{ \sum_i P(\mathcal{T}_a|\mathcal{T}_{o,i})P(\mathcal{T}_{o,i}| \beta,
\omega)}.
\end{equation*}
Once we have a trajectory in terms of options, the likelihood of taking that trajectory
is straightforward to compute given our model of the stochastic human policy:
\begin{equation*}
\label{eq:2} P(\mathcal{T}_{o,i} | \beta, \omega, \theta) = \prod_{k}\frac{\exp(\beta
Q^\odot( s_{ik}, o_{ik}))}{\sum_{o^\prime \in \omega} \exp(\beta Q^\odot(s_{ik}, o^\prime))},
\end{equation*}
where $o_{ik}$ denotes the option chosen in the $k^\mathrm{th}$ step of the $i^\mathrm{th}$ state-option trajectory, and $s_{ik}$ denotes the corresponding state.
To get the probability of the trajectory we multiply the probability of taking the individual option (given by our Boltzmann-rational model) across all options in the trajectory.
The likelihood for multiple observed trajectories follows
straightforwardly.

Procedure \ref{alg1} gives a method to compute all of the
option-trajectories which are consistent with a given action-trajectory. This
requires knowing the consistent-exit distribution \(P^{oc}(s_i,s_{i+k}, \mathcal{T}_a)\), as we need to know how likely activating an option is to give us the observed trajectory.

Since we have to enumerate each state-option trajectory \(\mathcal{T}_o\) which can
produce the observed state-action trajectory \(\mathcal{T}_a\), we should consider how
many of these state-option trajectories we may have. 
The Taxi-Driver case has a few `landmark' states which can be reached directly (via options) by many other states, while most states can only be reached by atomic actions from neighbouring states. If there are \(m\) of these landmark states which can each be reached by \(n\) other states, there are \(n^m\) possible option-trajectories consistent with the observed trajectory of actions. If we start introducing many states which can be destinations of landmarks, then the number of 
trajectories we have to consider increases exponentially. Of course, in principle humans can choose an arbitrary destination state for options, so in general the complexity of evaluating the BIHRL algorithm grows exponentially with the number of states in the problem. 

We could consider pruning the trees of
option-trajectories by removing any trajectories that have a very low probability as we create the sets of possible option-trajectories. However, this requires that we are very confident in our model of human behaviour in order to avoid removing trajectories that we erroneously think are unlikely. 

\begin{algorithm}[h!]
      \caption{Computing the full set of option-trajectories that are consistent with the observed state-action trajectory, and their corresponding probability. \\
    We successively step through the states in the observed trajectory. At each state we search to find all states that we can reach by triggering options in the current state. We form the list of all option-trajectories that can reach those states by concatenating the options that reach them with the list of options-trajectories that reach the current state.\\
    We sucessively update two sets: $\mathcal{T}_{oi}$ is the set of possible option sequences that account for the first $i$ actions, and $\mathcal{P}_{oi}$ are the corresponding probabilities that each sequence of options would produce the observed sequence of actions.}
  \begin{algorithmic}
    \REQUIRE 
        \begin{itemize}\addtolength{\itemsep}{-.6\baselineskip}
        \item{A computed optimal value function $V^\odot_B$ under a set of options $\omega$ with rationality parameter $\beta$}
        \item{ A function $P^{oc}(s_i, s_{i+k}, \mathcal{T}_{a, i:i+k})$ as defined in section \ref{sec:our_model}}
        \item{An observed state-action trajectory $\mathcal{T}_a$ of length $n$, with sub-trajectories between the \(i\) and \(i+k\) states $\mathcal{T}_{a, i:i+k}$}
    \end{itemize}
    \ENSURE A set of all trajectories of options that are consistent with the observed action-trajectory, along with the corresponding probabilities that taking that trajectory of options would result in the observed action-trajectory. \\
    \FOR{$i \in \{1,\dotsc, n\}$}
      \STATE $\mathcal{T}_{oi} \gets \emptyset$, $\mathcal{P}_{oi} \gets \emptyset$\\
    \ENDFOR
    \FOR{$i \in \{1, \dotsc, n\}$}
        \FOR{$k \in \{1, \dotsc, n - i\}$}
            \FOR{Each $o \in \omega$ with  $P^{oc}(s_i, \ s_{i+k}, \mathcal{T}_{a, i:i+k}) \neq 0$}
            \STATE  Generate a set of option-paths by appending all paths in         \(\mathcal{T}_{oi}\) with
            \(o\) and append these new option-paths to \(\mathcal{T}_{o(i+k)}\). \\Generate the corresponding probability by multiplying the probabilities in \(\mathcal{P}_{oi}\) by \(P^{oc}(s_i,s_{i+k}, \mathcal{T}_{a, i:i+k})\) and append these to \(\mathcal{P}_{o(i+k)}\).
            \ENDFOR
        \ENDFOR
    \ENDFOR
    \STATE \textbf{return} $\mathcal{T}_{on}, \mathcal{P}_{on}$
  \end{algorithmic}
    \label{alg1}
\end{algorithm}
\section{Taxi-Driver Experimental Results}
To illustrate how we carry out inference in this framework, we start by analysing our running example of the the taxi-driver environment. We use a simple MCMC method based on the
\texttt{Policy-Walk} algorithm from \citet{Ramachandran2007}, which
we describe in Appendix \ref{sec:appendix}. We use the family of reward
functions described in section \ref{sec:taxi-driver-description}, and
place a uniform prior over the number of cells that are free to
enter, running our method over five trajectories drawn from a
hierarchically-planning agent with a given true \(\theta\).
\begin{figure}
  \centering
  \includegraphics[width=0.5\textwidth]{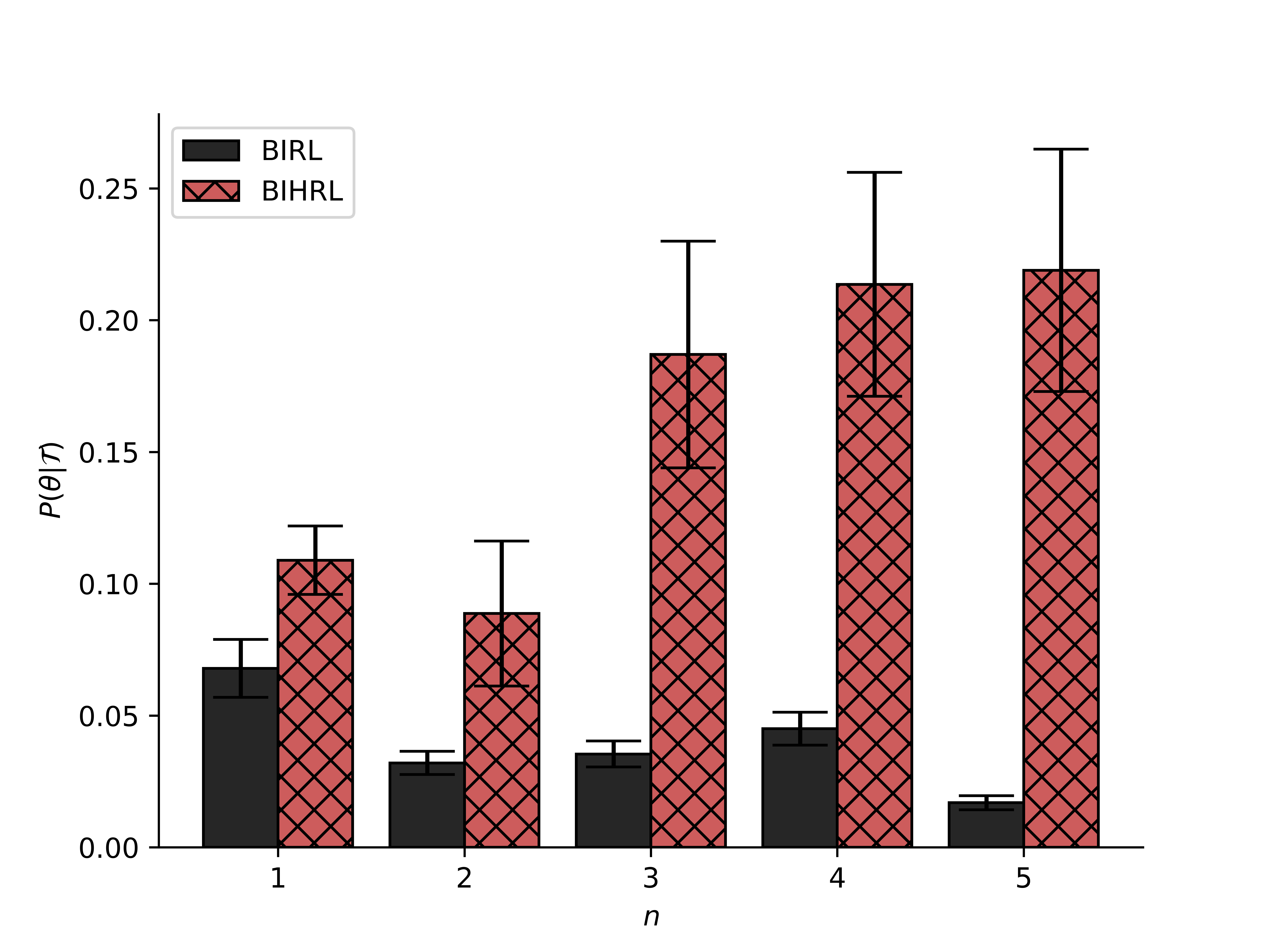}
  \caption{Bar chart showing the performance of the Bayesian IRL algorithm, with
    and without knowledge of hierarchical plans, at determining the true
    \(\theta\) from \(n\) trajectories. Error bars show one standard error in the mean over different MCMC seeds.}
  \label{fig:taxi-n}
\end{figure}
As we can see from the results in figure \ref{fig:taxi-n}, our knowledge of the
hierarchical structure of the agent's planning allows us to discern the true
\(\theta\) much better than assuming that the agent is merely a
self-consistent Boltzmann planner. We retain confidence in the true \(\theta\)
when seeing more and more trajectories, whilst the IRL algorithm without options
becomes increasingly convinced that the true \(\theta\) is not the correct
reward.

We can extend this simple example by analysing agents moving in much
more complicated environments, or by attempting to infer the option-sets that
the agents have available to them. We perform both in the following two
sections.

\section{Large-Scale Analysis: Wikispeedia}
Wikispeedia is an online game where players are given two
random articles from a subset of Wikipedia pages, and navigate from one page to
the other by clicking on hyperlinks, attempting to find the shortest path
from the first to the second. We apply our algorithm to a public dataset of thousands of Wikispeedia games, predicting the player's target Wikipedia page from the links traversed so far. This benchmark task has previously been studied by
\citet{West2012}. They hand-crafted a set of features, leaning heavily on the textual information in the pages to explain human
planning in the space. We apply our self-consistent hierarchical Boltzmann
planner to this task, to evaluate whether it can achieve comparable performance without
having to featurise the graph by hand.

This problem is conceptually similar to the taxi-driver problem, except that the
available actions are state-dependent, consisting of the hyperlinks that may be
clicked on each page. In the actual game, the players are able to click the
`back' button on the browser, which injects an additional action to consider. If
we were to include this action we would violate the Markov property of an MDP
(or complicate the analysis by squaring the size of the state space), so we only
consider those trajectories which don't use the back button. In order to
simplify our algorithm, we also ignore `dead-end' pages which don't link
anywhere. Finally, we removed paths longer than 20 steps long as they led to
computation difficulties and comprised less than 0.3\% of the dataset. We evenly
split the paths in the dataset into a training and testing set.

We model the player as an agent with uniform rewards of \(-1\) on all state
transitions except to the winning page, which delivers reward \(+20\). We postulate that humans may
choose long-time-scale strategies that attempt to navigate to specific pages in
particular.  Hence, we equip our agent with options that go to the \(m\) pages
that appear most frequently in the training set, with a common Boltzmann-rationality parameter \(\beta_o > \beta\). As an example, the top five
pages in the training set were \texttt{United States}, \texttt{Europe},
\texttt{United Kingdom}, \texttt{England}, and \texttt{Earth}.
\begin{figure} \centering
  \includegraphics[width=0.4\textwidth]{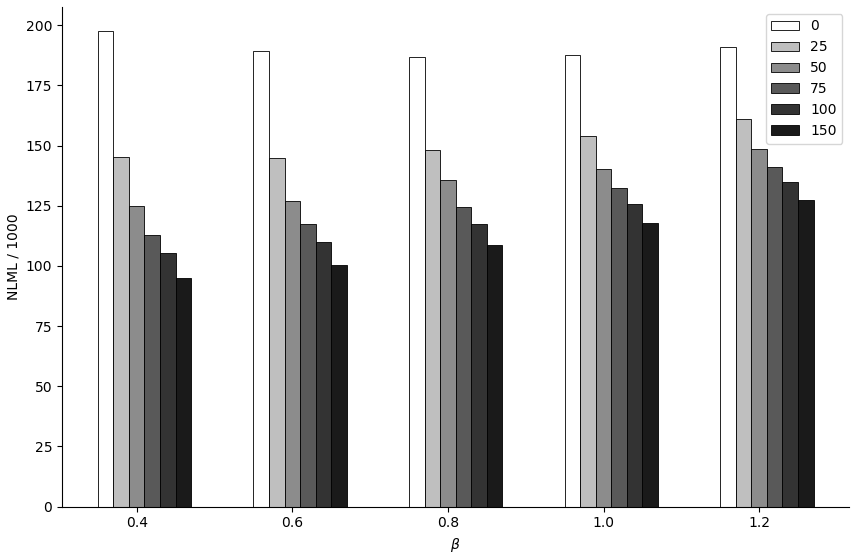}
  \caption{Showing the negative log marginal likelihood on the train set (lower
is better) for various combinations of the rationality constant \(\beta\), and
the number of hierarchical options \(m\), with darker bars corresponding to more
available options.The rationality of the options, \(\beta_o\), was fixed at 3.0.}
  \label{fig:wksp-NLML}
\end{figure}
With the choices made above, our agents are parameterised by the numbers \(m\),
\(\beta\), and \(\beta_o\).  We kept \(\beta_o\) fixed at 3.0 as initial
exploration showed little variation for different values as long as
they were substantially greater than \(\beta\). The discount rate \(\gamma\) was
fixed at 0.9. In order to find the the collection of hyperparameters \(\eta = (\beta, m)\), that best characterises the data, we compute the negative
log marginal likelihood (NLML), given by 
\begin{align*}
\mathrm{NLML} =& -\log(P(\{(\mathcal{T}_{a}, \theta)\}|\eta))\\ \propto &
-\log\left(\prod_iP(\mathcal{T}_{a,i}| \theta_i, \eta)\right)
\end{align*}
over all trajectories in the training set, and choose \(\eta\) such that the NLML is minimised. 

To compare our hierarchical planning model with \citet{West2012}, we consider trajectories
\(u_1, u_2, \ldots, u_n = \boldsymbol{u}_{1:n}\) consisting of \(n\) visited
articles \(u\), and observe the first \(k\) nodes. We then look at the
likelihood of predicting the correct target node compared to predicting another
node chosen uniformly at random from the nodes with the same shortest path
length from \(u_k\). This is given by
\begin{align}
  \frac{P(\theta| \boldsymbol{u}_{1:k}, \eta)}{P(\theta^\prime| \boldsymbol{u}_{1:k}, \eta)} &=
  \frac{P(\boldsymbol{u}_{1:k} | \theta, \eta)}{P(\boldsymbol{u}_{1:k} | \theta^\prime, \eta)}.
    \label{eq:evaluation}
\end{align}
We want to evaluate the ratio above for all of the data in the test set. Since the overwhelmingly most costly part of computing \(P(\boldsymbol{u}_{1:k} | \theta, \eta)\) is running the value iteration until convergence for each possible goal \(\theta\), we are able to speed up evaluation by precomputing the value functions beforehand.

\subsection{Results}
\begin{figure*}[!ht] \centering
  \includegraphics[width=0.9\textwidth]{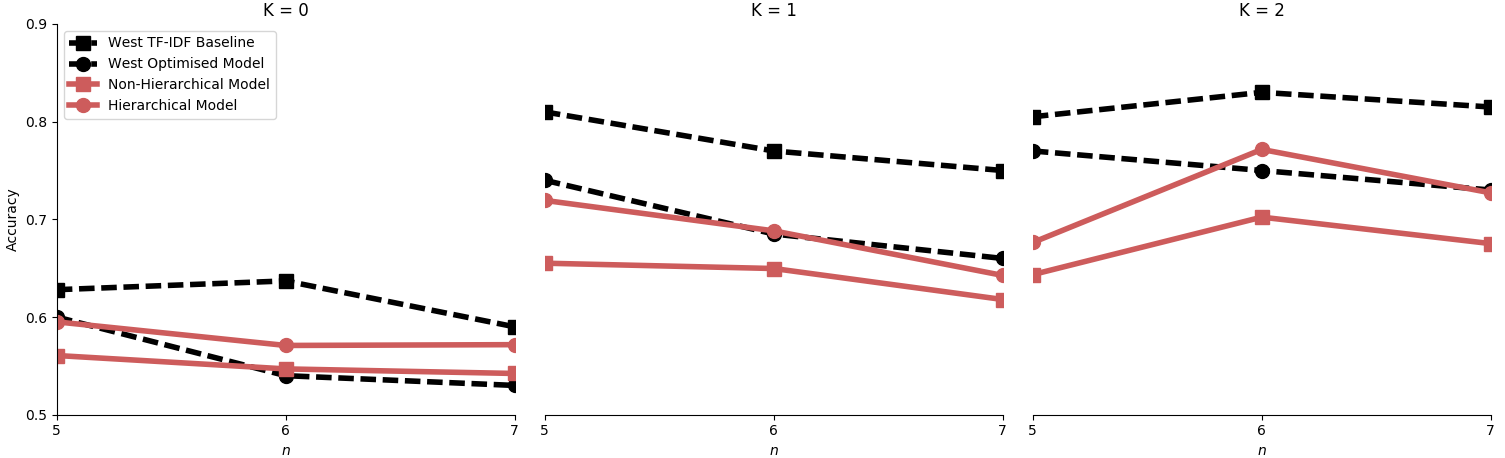}
  \caption{The accuracy on predicting \(\theta\) for a path of length \(n\)
    given the first \(k\) nodes. }
  \label{fig:wksp-results}
\end{figure*}%
Figure \ref{fig:wksp-NLML} shows that including a set
of hierarchical options decreases the NLML by a factor of two. When our agents
have no hierarchical actions, changing \(\beta\) has a negligible effect on the NLML.
We also observe that the minimal NLML is obtained with a large set
of around 150 available hierarchical options. It seems reasonable to us that a typical
player may know one or two hundred topics well enough to navigate expertly to
them (with \(\beta_0 = 3.0\)), whilst the other randomly drawn topics are not
known well at all (with \(\beta = 0.4\)).

Figure \ref{fig:wksp-results} shows the predictive performance of our
hierarchical model. We note that including hierarchical policies provides a
substantial benefit over the BIRL baseline, taking the accuracy
from an average of 62\% to 66\%. The model with hierarchical policies
performs comparably to \citet{West2012}'s TF-IDF algorithm based on semantic
similarity of topics, although we remain below the state-of-the-art results
obtained by their hand-crafted featurisation. 
\section{Inferring Option-Sets}
\label{sec:priors}
If we don't know the options available to the human, we might want to
infer what those are, and marginalise over these, i.e.\ compute
\begin{equation*} P(\theta | \beta, \mathcal{T}_a) = \int_\Omega P(\theta
|\mathcal{T}_a, \beta, \omega)P(\omega)d \omega,
\end{equation*} integrating over all sets of options \(\omega\) in the space of
possible sets of options \(\Omega\).  In general, there are a very large number
of possible options. Even simply considering deterministic options, there are
\(|\mathcal{S}|^\mathcal{|A|}\) possible options, and the set of all possible
sets of options is exponentially larger again: \(|\Omega| =
2^{\mathcal{|S|}^\mathcal{|A|}}\).

Given the large size of the latent space, marginalising over all option-sets to infer the posterior distribution
over \(\theta\) quickly becomes computationally intractable. Future work could try to tame this
intractability by utilising recent advances in Hamiltonian Monte-Carlo
approaches and variational inference. Here, we tackle the simpler case
of the taxi-driver with the naive MCMC approach to show that this
approach can learn interesting results.

We equip the MCMC method with a prior over \(\Omega\) which is uniform over all
sets of up to three options, with each option consisting of a deterministic
policy that executes direction steps in order to optimally navigate to a given
destination which is chosen from a set of \(16\) cells which are close to the landmarks
and shown in figure \ref{fig:prob2}. Note that this excludes the \(9\) cells in the
middle of the grid which aren't close to any destinations. This captures the
skills we would expect a driver to use in the environment, with skills that go to
the areas of the grid that are near the landmarks where the passengers are picked up
and put down. We keep our prior over \(\theta\) as before.

\begin{figure}[htb]
  \includegraphics[width=0.5\textwidth]{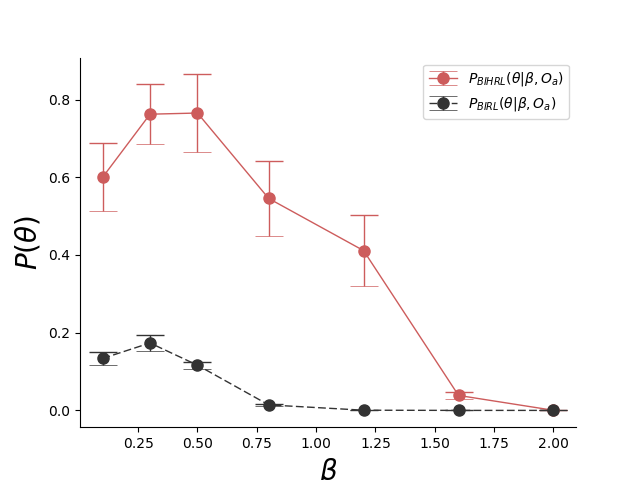}
  \caption{Probabilities assigned to \(\theta_0\), the ground truth reward,
    when conditioned on five trajectories from a
    hierarchical planner with \(\beta = 0.8\), marginalising over the
    option-sets described in the text.}
  \label{fig:results}
\end{figure}
\subsection{Results}

The results in figure \ref{fig:results} show that even if we do not know the
options used to plan, but merely have a prior distribution over them, BIHRL
predicts the ground truth reward \(\theta_0\) with higher
probability than BIRL. BIRL predicts a probability of less than 0.03 and BIHRL a
probability 0.55 at the ground-truth \(\beta\). 

This experiment demonstrates that the BIHRL model is able to infer the preferences
from the actions of hierarchical planners, without necessarily knowing the options
a priori. However, our naive MCMC method will not scale to substantially larger
latent state spaces, such as the space of 150 latent options that would be required to
extend this to the Wikispeedia dataset.
\section{Conclusion}
We have extended inverse reinforcement learning to infer preferences
from hierarchical planners which choose among options with a
self-consistent Boltzmann-policy.  We show that these agents capture
many of the tradeoffs between the reward and the cost of gathering
information that humans intuitively make.

We introduce an inference algorithm based on the \texttt{Policy-Walk} algorithm
developed by \citet{Ramachandran2007} and show that it infers preferences of
hierarchical planners much more accurately than standard Bayesian IRL on an
illustrative toy example based on the taxi-driver environment from
\citet{Dietterich2000}. Further, including a straightforward set of hierarchical
plans significantly increases the accuracy of modelled human planning in the
`Wikispeedia' dataset introduced by \citet{West2012}, taking the accuracy
from an average of 62\% to 66\%. Our method obtains comparable
accuracy to the baseline of \citet{West2012}, despite not relying
on any hand-engineered features.

We discussed how we would deal with the case where we do not know our
planners' hierarchical options a priori, and are forced to infer
agents' available options jointly along with the reward. We introduce
a toy MCMC approach that is able to infer the correct option-sets and
reward for small environments. Given the correct \(\beta\), BIHRL
assigns 20 times more probability mass to the ground-truth \(\theta\)
than standard BIRL. 

However, at present significant challenges remain for using BIHRL in practical environments, consisting of long trajectories of agents with complex options. The large number of possible options that realistic planners could use means that any inference procedure must deal with very high-dimensional probability distributions, while the relative complexity of actual human options means that it is computationally intractable to generate the exponential numbers of plausible option-trajectories that are consistent with the observed action-trajectory. It is possible that very good models of human behaviour may be able to cut down the exponential numbers of human choices, by assigning strong priors over which human behvaiors and actions are likely. Furthermore, modern Hamiltonian MC and variational inference may be able to assist with the inference in high-dimensional spaces. If we can solve these daunting problems, we may be able to use BIHRL to more accurately infer human preferences in a variety of
complicated situations.





\appendix
\vspace{1cm}
\section{MCMC Sampling Procedure}
\label{sec:appendix}
    

         

\begin{algorithm}
  \caption{MCMC sampling in the latent space of \(\Theta, \Omega\).}
\begin{algorithmic}
      \REQUIRE{\begin{itemize}
        \item{A set of possible reward functions \(\Theta\)}
        \item{A set of possible options \(\Omega\)}
        \item{A set of trajectories \(\mathcal{T}_{a,i}\)}
      \end{itemize}}
      \ENSURE{Samples from the posterior distribution \(P(\theta, \omega | \mathcal{T}_a, \beta)\)}
         \STATE {$p \gets 0.5$}
         \STATE {$V \gets 0$}
         \STATE {$\theta \gets \mathrm{Random \ Draw}(\Theta)$}
         \STATE {$\omega \gets \mathrm{Random \ Draw}(\Omega)$}
         \STATE Samples \(\gets\) Empty list
         \REPEAT
             \STATE {Pick \(\theta_1\) and \(\omega_1\) randomly amongst the neighbours of \(\theta\), \(\omega\)}
             \STATE {$V_1 \gets \mathrm{Value \ Iteration}(\beta, \omega_1, \theta_1$), where the value iteration is initialised with $V$}
             \STATE {Compute \(p_1 = P(\mathcal{T}_a| \beta, \omega_1, \theta_1) \times P(\omega_1 | \Omega)\)}
             \STATE {With probability \(\min(1, p_1/p)\):{\\ \qquad \(p \gets p_1\) \\ \qquad \(V \gets V_1\) \\ \qquad \(\theta \gets \theta_1\) \\ \qquad \(\omega \gets \omega_1\)\\}}
        Append \((\theta, \omega)\) to Samples
         \UNTIL{Desired number of samples obtained}
         \STATE \textbf{return} Samples
\end{algorithmic}    
\end{algorithm}

\end{document}